\documentclass[10pt,twocolumn,letterpaper]{article}

\usepackage{cvpr}              

\newcommand{\norm}[1]{\left\lVert#1\right\rVert}

\usepackage{graphicx}
\usepackage{algorithm}
\usepackage{algpseudocode}
\usepackage{comment}
\usepackage{graphicx}
\usepackage{booktabs}
\usepackage{pifont}
\usepackage{xcolor}
\definecolor{mygreen}{HTML}{196B24}
\definecolor{myred}{HTML}{C00000}
\usepackage[margin=1in]{geometry}

\usepackage{caption} 
\definecolor{cvprblue}{rgb}{0.21,0.49,0.74}
\usepackage[pagebackref,breaklinks,colorlinks,allcolors=cvprblue]{hyperref}

\def\paperID{2037} 
\def\confName{CVPR}
\def\confYear{2025}

\title{VENOM: Text-driven Unrestricted Adversarial Example Generation with Diffusion Models}

\author{Hui Kuurila-Zhang, Haoyu chen, Guoying Zhao\thanks{Corresponding author. Email: guoying.zhao@oulu.fi}\\
University of Oulu, CMVS\\
{\tt\small \{hui.zhang, chen.haoyu, guoying.zhao\}@oulu.fi}}

\begin{document}

\twocolumn[{
\maketitle
\vspace{-30pt}
\begin{center}
    \captionsetup{type=figure}
    \includegraphics[width=\textwidth]{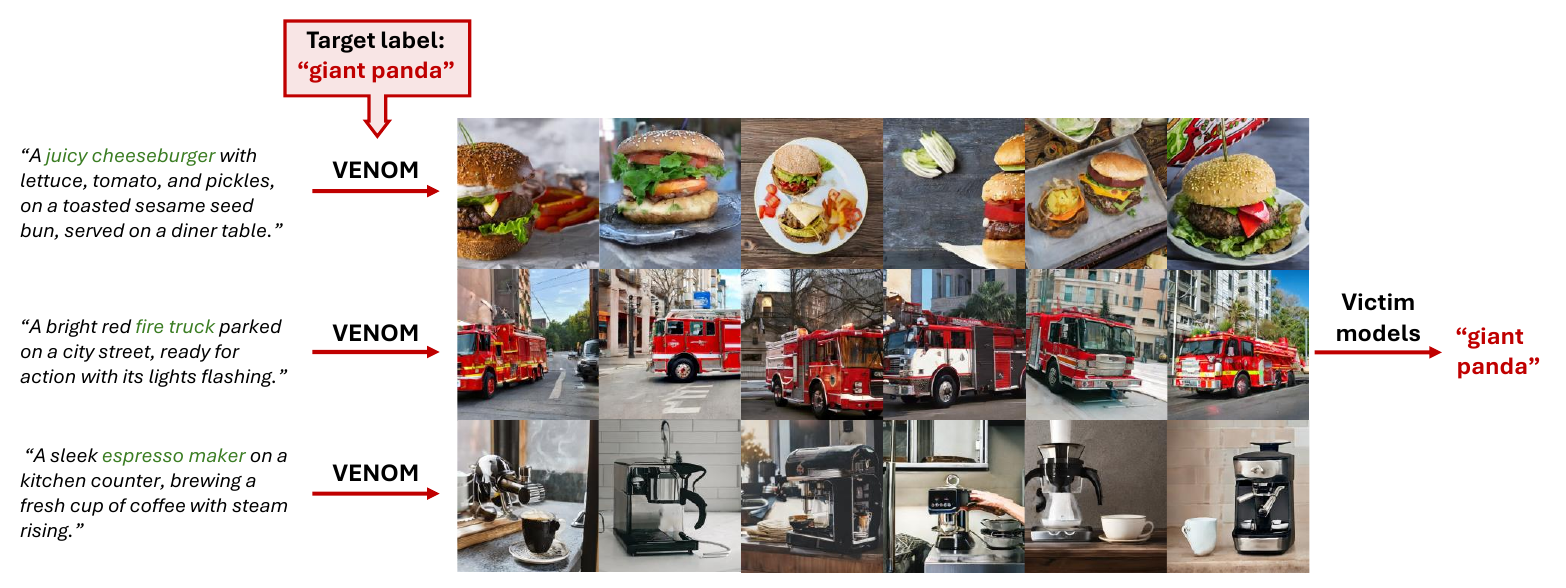}
    \captionof{figure}{Samples of Natural Adversarial Examples (NAEs) generated by VENOM, conditioned on input text prompts and a designated target label. VENOM achieves nearly a 100\% white-box attack success rate in generating NAEs while preserving high image fidelity.}
    \label{fig:firstpageexamples}
\end{center}
}]

\renewcommand\thefootnote{\fnsymbol{footnote}}
\footnotetext[1]{Corresponding author. Email: guoying.zhao@oulu.fi}
\renewcommand\thefootnote{\arabic{footnote}}


\begin{abstract}

\vspace{-4pt}
\noindent Adversarial attacks have proven effective in deceiving machine learning models by subtly altering input images, motivating extensive research in recent years. Traditional methods constrain perturbations within $l_p$-norm bounds, but advancements in Unrestricted Adversarial Examples (UAEs) allow for more complex, generative-model-based manipulations. Diffusion models now lead UAE generation due to superior stability and image quality over GANs. However, existing diffusion-based UAE methods are limited to using reference images and face challenges in generating Natural Adversarial Examples (NAEs) directly from random noise, often producing uncontrolled or distorted outputs.

In this work, we introduce \textbf{VENOM}, the first text-driven framework for high-quality unrestricted ad\textbf{V}ersarial \textbf{E}xamples ge\textbf{N}eration through diffusi\textbf{O}n \textbf{M}odels. VENOM unifies image content generation and adversarial synthesis into a single reverse diffusion process, enabling high-fidelity adversarial examples without sacrificing attack success rate (ASR). To stabilize this process, we incorporate an adaptive adversarial guidance strategy with momentum, ensuring that the generated adversarial examples $x^*$ align with the distribution $p(x)$ of natural images. Extensive experiments demonstrate that VENOM achieves superior ASR and image quality compared to prior methods, marking a significant advancement in adversarial example generation and providing insights into model vulnerabilities for improved defense development.

\end{abstract}    
\vspace{-10pt}
\section{Introduction}
\label{sec:intro}

The increasing sophistication of deep learning models has brought about a parallel rise in the need to understand their vulnerabilities, especially through adversarial attacks, where intentionally perturbed inputs, known as adversarial examples \cite{AEs}, are crafted to deceive the victim models with high confidence. These adversarial examples pose a significant threat to the robustness and reliability of deep learning systems. Traditional adversarial attack methods \cite{fgsm, madry_at, cw_attack, autoattack} generate adversarial examples by perturbing clean inputs within a restricted $l_p$-norm ball of magnitude $\epsilon$, aiming to deceive the victim models while maintaining the imperceptibility. However, more recent research has introduced Unrestricted Adversarial Examples (UAEs) \cite{uae_origin}, which are generated without constraints on perturbation magnitude. UAEs can be synthesized directly from random noise using generative models, allowing for more complex transformations beyond pixel-level perturbations.

\noindent UAEs have demonstrated enhanced attack success rates \cite{diffattack, advdiff}, especially against robust defense methods, compared to traditional, perturbation-based adversarial examples. Initially, Generative Adversarial Networks (GANs) were widely used for generating UAEs \cite{ganuae1,ganuae2,uae_origin}, but recent diffusion models have outperformed GANs in image quality due to their more stable training process. Diffusion models, by refining the process of image generation, have become the preferred choice for synthesizing UAEs in recent works \cite{advdiff, sd-nae, diffattack, advdiffuser}, all of which demonstrated superior performance over GAN-based methods. Importantly, these diffusion-based methods do not require re-training diffusion models for adversarial purposes; rather, they perturb the reverse diffusion process of pre-trained diffusion models to generate UAEs. However, because pre-trained diffusion models are optimized for generating natural images, adversarial guidance during the reverse sampling process can destabilize the generation, resulting in corrupted or unnatural images that may fall outside the distribution of realistic images.

In practice, injecting adversarial guidance into the reverse diffusion process without precise control has led existing diffusion-based UAE generation methods \cite{advdiff, sd-nae, diffattack, advdiffuser} to produce inconsistent results. For instance, \cite{sd-nae} often generates adversarial examples that are perceptually invalid, while \cite{advdiff} produces overly perturbed images due to uncontrolled adversarial gradients, \cite{advdiffuser} suffers from under- or over-perturbed adversarial outputs. 
While \cite{diffattack} demonstrates stability in generating UAEs from reference images, it is unable to produce Natural Adversarial Examples (NAEs)—a subset of UAEs generated directly from random noise. Hereafter, we use UAE to refer to both UAE and NAE, reserving NAE specifically when emphasizing generation from random noise. Figure \ref{fig:experiment1} illustrates how our proposed method surpasses existing NAE generation approaches \cite{advdiff, advdiffuser, sd-nae} in terms of image quality and consistency. The limitations of these prior methods highlight the necessity for a stable, controlled mechanism to generate high-quality UAEs and NAEs that preserve adversarial effectiveness without sacrificing visual realism. 

\noindent To address these challenges, we propose VENOM, a novel framework for stable and high-quality UAE and NAE generation. VENOM introduces an adaptive control strategy to control the injection of gradient-based adversarial guidance, minimizing the risk of image corruption while maintaining fidelity to the natural image distribution. Additionally, we incorporate a momentum-based technique into the adversarial gradient to enhance transferability and attack strength. Leveraging the Stable Diffusion model \cite{stable_diffusion}, along with our adaptive control and momentum modules, VENOM enables text-driven, UAE generation. As the first framework to support fully text-driven UAE generation, VENOM allows users to specify both the visual content and target class of the generated adversarial examples, providing flexibility in crafting adversarial images entirely based on user-defined prompts (see Figure \ref{fig:firstpageexamples}).

VENOM achieves near 100\% attack success rate (ASR) on white-box models and maintains strong ASR against defense methods while preserving high image fidelity. The framework offers further versatility by supporting both UAE and NAE modes: adversarial examples generated from random noise are classified as NAEs, while those created by perturbing reference images are designated as UAEs.

Our work makes the following key contributions: 
\begin{itemize} 
\item We present VENOM, the first framework, to our knowledge, for text-driven UAEs generation, enabling customized image content and adversarial attack generation purely through text prompts.
\item VENOM introduces the first fully-integrated pipeline that combines image content generation and adversarial attack synthesis in a unified reverse diffusion process.
\item We propose an adaptive control strategy that precisely modulates adversarial guidance to ensure high image quality with minimal artifacts or distortions. Additionally, we incorporate a gradient-based adversarial guidance mechanism with momentum, compensating for any potential reduction in attack strength from the adaptive control strategy. This approach enables our framework to achieve a balance between image fidelity and adversarial robustness.
\item Our framework offers high adaptability, supporting both UAE and NAE modes: generating images from random noise in NAE mode and perturbing reference images in UAE mode.
\end{itemize}

By addressing the limitations of existing diffusion-based UAE generation methods, VENOM not only advances the state of the art in adversarial example generation but also provides a robust and flexible tool for studying model vulnerabilities. Our approach holds promise for both advancing adversarial robustness research and developing new insights into the potential weaknesses of deep learning systems.

\section{Related Work}
\label{sec:relate_work}
\subsection{Adversarial Examples}

The concept of Adversarial Examples (AEs), first introduced by \cite{AEs}, denotes maliciously crafted data generated by applying a small perturbation, $\delta$, to clean data $x$. This alteration yields a perturbed instance $x^*$, designed to mislead the victim models into incorrect predictions. Formally, an adversarial example is defined as:
\begin{equation} x^* = x + \delta \quad \text{s.t.} \quad f(x^*) \neq y \quad \text{and} \quad \lVert \delta \rVert_p < \epsilon \end{equation}
where $f$ denotes the victim model, $y$ represents the true label, and $\epsilon$ bounds the perturbation’s magnitude within an $l_p$ norm to ensure perceptual similarity to $x$. A variety of algorithms—such as FGSM \cite{fgsm}, PGD \cite{madry_at}, CW \cite{cw_attack}, and AutoAttack \cite{autoattack}—have been developed to construct AEs under this constraint.

\subsection{Unrestricted Adversarial Examples}
Unrestricted Adversarial Examples (UAEs), as the name suggests, extend beyond traditional $\epsilon$-bounded perturbations by allowing adversarial examples without the constraint of small, imperceptible changes. UAEs are defined within the set $\mathcal{A}$:
\begin{equation} 
\mathcal{A} \triangleq \{x \in \mathcal{O} \mid o(x) \neq f(x)\} 
\label{eq:defuae}
\end{equation}
where $\mathcal{O}$ denotes the set of images perceived as natural and realistic by humans, $o(x)$ represents human evaluation of the image $x$, and $f(x)$ is the prediction from the victim model. This broader definition permits UAEs to involve transformations such as rotations \cite{uaebrown}, texture alterations \cite{uaetexture, uaetexture2}, or entirely synthesized images \cite{uae_origin}. Generative Adversarial Networks (GANs) have been commonly employed to synthesize UAEs from scratch \cite{ganuae1, ganuae2, uaenae, uae_origin}. However, recent diffusion model-based methods for generating UAEs \cite{advdiffuser, sd-nae, advdiff, diffattack} have highlighted limitations of GAN-based attacks, which suffer from poor image quality and low attack success rates (ASR), especially on large-scale datasets like ImageNet \cite{imagenet}. Specifically, \cite{sd-nae} synthesizes UAEs by perturbing text embeddings within Stable Diffusion models, resulting in less distortion. \cite{diffattack} optimizes latent representations by maximizing the likelihood of misclassification. \cite{advdiffuser} incorporates PGD attack into the reverse diffusion process, while \cite{advdiff} proposes adversarial sampling during the reverse diffusion phase.

\subsection{Natural Adversarial Examples}
Natural Adversarial Examples (NAEs) are adversarial instances arising from naturally occurring variations in data, as opposed to artificial perturbations, and are defined in Eq. (\ref{eq:defuae}). Originally introduced by \cite{nae_origin}, NAEs represent real-world, out-of-distribution examples that challenge deep learning models due to distributional shifts \cite{naeweather, naeweather2}, such as variations in lighting, viewpoint, or environmental conditions.

We consider NAEs to be a subset of UAEs, as UAEs encompass both naturally occurring adversarial instances and artificially perturbed cases. In some studies, the terms NAEs and UAEs are used interchangeably \cite{advdiffuser, sd-nae}, though NAEs strictly represent a narrower subset of UAEs. In our work, we designate UAEs generated from random noise as NAEs, while UAEs generated by perturbing existing images are simply referred to as UAEs.

\section{Method}
\label{sec:method}
%
\begin{figure*}[ht]
    \centering
    \includegraphics[width=0.9\linewidth]{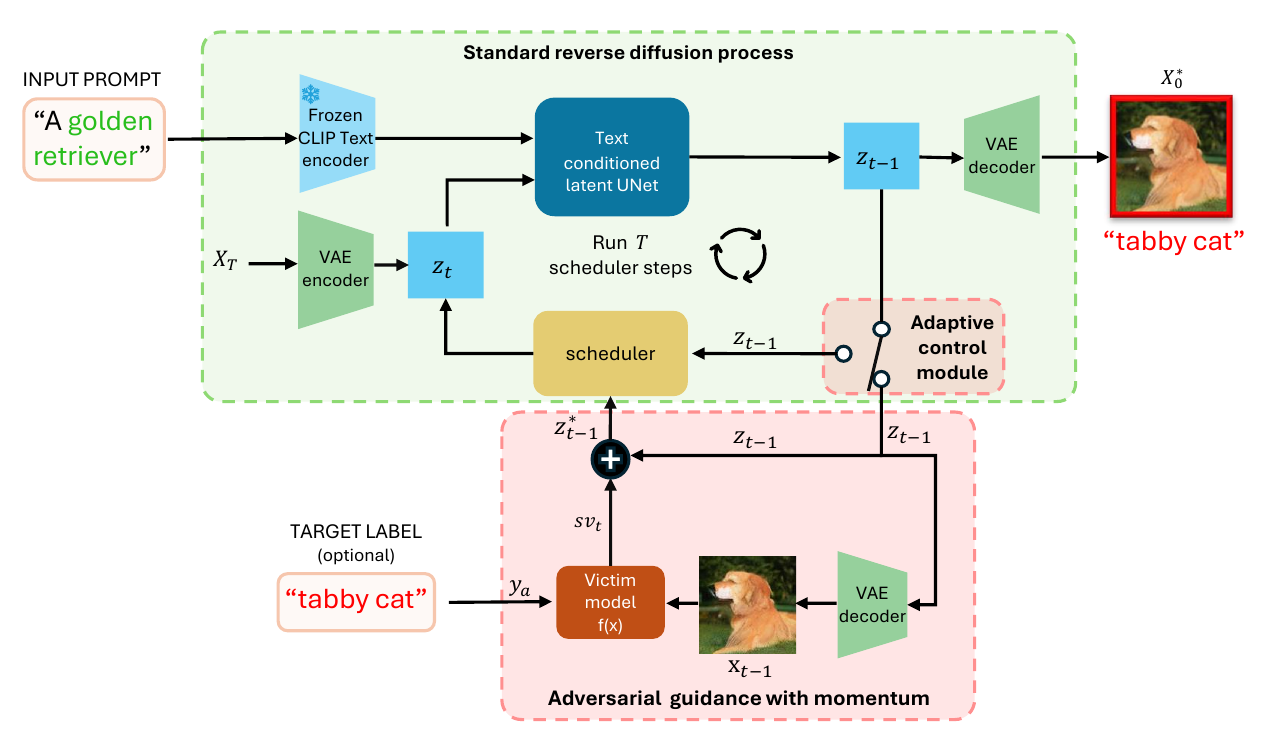}
    \caption{The overview of VENOM algorithm for generating NAEs (no reference images) and UAEs (with reference images). In NAE model, the input $X_T$ is sampled from the standard Gaussian distribution $ \mathcal{N}(0,1)$. In UAE mode, the input $X_T$ is derived by applying the DDIM inversion (Eq. (\ref{eq:ddim_inversion})) to the reference image $X_0$. If the target label is unavailable, the class with the second-highest likelihood, excluding the ground truth, is assigned as the target label.}
    \label{fig:VENOM_fig}
\end{figure*}

As discussed in Section \ref{sec:relate_work}, previous adversarial attack methods based on diffusion models \cite{advdiff,advdiffuser,sd-nae,diffattack} face a trade-off between the Attack Success Rate (ASR), and image quality. To address this limitation and achieve both high ASR and high image quality, we propose a novel framework designed to stabilize adversarial guidance during the reverse diffusion process by using an adaptive control strategy and momentum, which we term VENOM. The adaptive control strategy and momentum are designed to guide the stable reverse diffusion process to generate adversarial examples, denoted as $x^*$, from the same distribution $p(x)$ as natural images, ensuring that $x^*$ functions as both a natural and adversarial image simultaneously.

Figure \ref{fig:VENOM_fig} depicts the pipeline for generating adversarial examples from either random noise or reference images using the VENOM algorithm. VENOM offers high flexibility, allowing the generation of adversarial examples based solely on text prompts, with both the target label and reference image being optional. The pipeline is composed of three key modules: the standard reverse diffusion process of the stable diffusion model, adversarial guidance with momentum, and an adaptive control strategy for integrating the adversarial guidance. The following subsections provide a detailed explanation of each module.
\subsection{Stable Diffusion}
Stable Diffusion \cite{stable_diffusion} is a latent text-to-image diffusion model that incrementally refines random noise into high-quality images guided by text prompts. The approach is rooted in the Denoising Diffusion Probabilistic Model (DDPM) \cite{ddpm}, consisting of a forward diffusion phase, where noise is successively added to a clean input $z_0$ over $T$ discrete time steps, and a reverse phase, aiming to reconstruct the input from a noisy latent $z_T$. 
\begin{equation}
    q(z_t|z_{t-1}) = \mathcal{N}(z_t;\sqrt{1-\beta_t}(z_{t-1}), \beta_t\textbf{\textrm{I}}),
    \label{eq:forward1}
\end{equation}
Eq. (\ref{eq:forward1}) describes the forward diffusion process, where $\beta_t\in(0,1)$ is a variance scheduler. $z_T$ is equivalent to an isotropic Gaussian distribution when $T \rightarrow \infty$. 
By leveraging the properties of Gaussian distributions, the noisy latent at any time step $t$ can be sampled directly from $z_0$ as shown in Eq. (\ref{eq:forwar2}), where $\Bar{\alpha}_t=\Pi^t_{s=1}(1-\beta_s)$.
\begin{equation}
    q(z_t|z_0) = \mathcal{N}(z_t;\sqrt{\Bar{\alpha}_t}, (1-\Bar{\alpha}_t)\textbf{\textrm{I}}),
    \label{eq:forwar2}
\end{equation}

The diffusion model $\epsilon_\theta(z_t, t)$ predicts the noise added to $z_t$ at step $t$, enabling denoising from $z_t$ to $z_{t-1}$ by removing the predicted noise. The model $\epsilon_\theta$ is trained by minimizing the following loss function\cite{ddpm}:
\begin{equation}
   \underset{\theta}{\mathrm{min}}\mathcal{L}(\theta) = \mathbb{E}_{z_0, \epsilon \sim N(0,I),t}\norm{\epsilon-\epsilon_\theta(z_t,t)}^2_2,
\end{equation}

The DDPM \cite{ddpm} reverse process is a Markov chain of stochastic Gaussian transitions requiring a substantial number of time steps $T$, which results in prolonged inference. To expedite this process while maintaining sample quality, \cite{ddim} introduced the Denoising Diffusion Implicit Model (DDIM), offering faster inference with fewer diffusion steps and a deterministic reverse process. Therefore, DDIM is utilized as the scheduler in the Stable Diffusion pipeline in our project. The reverse process under DDIM is defined in Eq. (\ref{eq:ddim_reverse}).
\begin{equation}
    z_{t-1} = \sqrt{\Bar{\alpha}_{t-1}}(\frac{z_t - \sqrt{1-\Bar{\alpha}_t}\epsilon_\theta}{\sqrt{\Bar{\alpha}_t}}) + \sqrt{1-\Bar{\alpha}_{t-1}}\epsilon_\theta), 
    \label{eq:ddim_reverse}
\end{equation}

VENOM generates UAE from the latent noisy input $z_T$, which may originate from a Gaussian noise $\mathcal{N}(0,1)$ or a noisy latent derived from a reference image after forward diffusion. Benefiting from DDIM's deterministic property \cite{ddim}, we can invert the reverse process by Eq. (\ref{eq:ddim_inversion}) to encode the reference image into a chosen noisy latent $z_{t+1}$ at any step $t\in[0,1,...,T-1]$, rather than executing forward diffusion from $z_0$. Note that $\epsilon_\theta$ in Eq. (\ref{eq:ddim_reverse}) and Eq. (\ref{eq:ddim_inversion}) is a simplified notation for\ $\epsilon_\theta(z_t,t)$.
\begin{equation}
    z_{t+1} = \sqrt{\Bar{\alpha}_{t+1}}(\frac{z_t - \sqrt{1-\Bar{\alpha}_t}\epsilon_\theta}{\sqrt{\Bar{\alpha}_t}}) + \sqrt{1-\Bar{\alpha}_{t+1}}\epsilon_\theta),
    \label{eq:ddim_inversion}
\end{equation}
\subsection{Adversarial Guidance with Momentum}
The adversarial guidance in VENOM is applied to the latent variable $z_{t-1}$ following each denoising step ($z_t\rightarrow z_{t-1}$) when adversarial perturbation is activated; otherwise, $z_{t-1}$ proceeds to the next denoising step without modification, as shown in Figure \ref{fig:VENOM_fig}. Unlike most diffusion-based adversarial attack approaches, which iteratively optimize the latent variable $z_t$ or input embeddings \cite{diffattack, advdiffuser, sd-nae}, we apply a small adversarial perturbation incrementally at each reverse step, as demonstrated in \cite{advdiff}. This approach reduces visual distortion and enhances the image quality of the generated UAEs.

\noindent The adversarial guidance at each reverse diffusion step is similar to FGSM \cite{fgsm} but differs in that it uses an unrestricted gradient without any $l_p$-norm constraints. Instead of relying solely on the instantaneous gradient at each step, we introduce a momentum-based approach, where a moving average of past gradients stabilizes the adversarial guidance. This technique allows for consistent adversarial directionality without compromising image quality, effectively guiding the generation towards adversarial examples while maintaining visual fidelity.
At each reverse diffusion step $t$, the adversarial guidance $g(t)$ is calculated as follows:
\begin{equation}
    g(t) = \nabla_{x_{t-1}}\textrm{log}p_f(y_a|x_{t-1})
    \label{eq:compute-gradient}
\end{equation}
where $y_a$ is the target label that the adversarial example is intended to misclassify, and $x_{t-1}$ represents the one-step denoised version of $x_t$ during the reverse diffusion process.
\noindent The adversarial guidance is then incorporated into the reverse diffusion process with momentum, as shown in Eq. (\ref{eq:adv-sampling}):
\begin{equation}
    z^*_{t-1} = z_{t-1} + sv_{t}:\begin{cases}
        v_{t_0} = g(t)& t = t_0\\
        \beta v_{t + 1} + (1-\beta) g(t) & t < t_0
    \end{cases}
    \label{eq:adv-sampling}
\end{equation}

Here, $v_t$ represents the momentum-adjusted adversarial guidance, scaled by a factor $s$. At the initial step $t_0$, $v_t$ is initialized as $g(t_0)$. In subsequent steps, it is computed as an exponentially weighted moving average of previous gradients, controlled by the momentum coefficient $\beta$. This momentum-based adversarial guidance smooths the perturbations, reducing the risk of image corruption while ensuring the generation remains adversarially effective.

\begin{algorithm}[ht]
\caption{The Algorithm of VENOM}\label{alg:VENOM}
\begin{algorithmic}

\Require $T$: diffusion time steps; $f$: victim model; $N$: attack iterations; $x$: (optional) reference image; $y_a$: (optional) target label; $t_{start}$: start step for adversarial guidance

\State $x_T \gets \texttt{DDIM\_inversion}(x) \textbf{ if } X \text{ exists else } \mathcal{N}(0,1)$
\State $z_T \gets \texttt{VAE\_encoder}(x_T)$
\State $\texttt{adv\_guidance} \gets \texttt{True}$

\For{$N \in [1, 2, ..., N]$}
    \For{$t \in [T, T-1, ..., 0]$}
        \If{$N > 2$} $\texttt{adv\_guidance} \gets \texttt{True}$ \EndIf
        \State $z_{t-1} \gets \texttt{reverse\_diffusion\_step}(z_t)$

        \If{$\texttt{adv\_guidance} == \texttt{False}$}
            \State $x_{t-1} \gets \texttt{VAE\_decoder}(z_{t-1})$
            \If{$y_a \neq \text{argmax}f(x_{t-1})$} $\texttt{adv\_guidance} \gets \texttt{True}$ \EndIf
        \EndIf

        \If{$0 < t \leq t_{start} \textbf{ and } \texttt{adv\_guidance}$}
            \State $x_{t-1} \gets \texttt{VAE\_decoder}(z_{t-1})$
            \If{$y_a == \text{argmax}f(x_{t-1})$} $\texttt{adv\_guidance} \gets \texttt{False}$ \EndIf
            \If{$t = t_{start}$} \State $v\gets \nabla_{x_{t-1}} \texttt{log}p_f(y_a|x_{t-1})$ \EndIf
            \State $v \gets \beta v + (1-\beta)\nabla_{x_{t-1}}\texttt{log}p_f(y_a|x_{t-1})$
            \State $z_{t-1} \gets z_{t-1} + s v$
        \EndIf
    \EndFor
    \State $x_0 \gets \texttt{VAE\_decoder}(z_0)$
    \If{$y_a == \text{argmax}f(x_0)$} \textbf{break} \EndIf
\EndFor

\State \Return $x_0$
\end{algorithmic}
\end{algorithm}

\subsection{Adaptive Control Strategy for Injecting the Adversarial Guidance}
As illustrated in Figure \ref{fig:VENOM_fig}, the adaptive control strategy functions as a switch that governs when to activate or deactivate adversarial guidance. This straightforward switching mechanism effectively adjusts the strength of the adversarial attack in response to the current state of the latent variable $z_t$. The detailed procedure is outlined in Algorithm \ref{alg:VENOM}. The adversarial guidance is toggled according to the following conditions:
\begin{enumerate}

    \item  \textbf{ON}: By default.
    \item  \textbf{OFF}: If $x_t$ is successfully transformed in to an adversarial example.
    \item  \textbf{ON}: if adversarial guidance is detected as OFF, but $x_t$ is no longer adversarial after subsequent denoising.
    \item \textbf{ON}: If the attack fails twice, adversarial guidance is forcefully kept ON in future iterations.
\end{enumerate}

This adaptive yet simple control strategy enables VENOM to generate UAEs with a high ASR while maintaining superior image quality. To the best of our knowledge, VENOM is the first UAE generation method that dynamically controls attack strength and achieves the best image quality among existing approaches.

\vspace{-4pt}
\section{Experiments}
\begin{figure*}[ht]
    \centering
    \includegraphics[width=\linewidth]{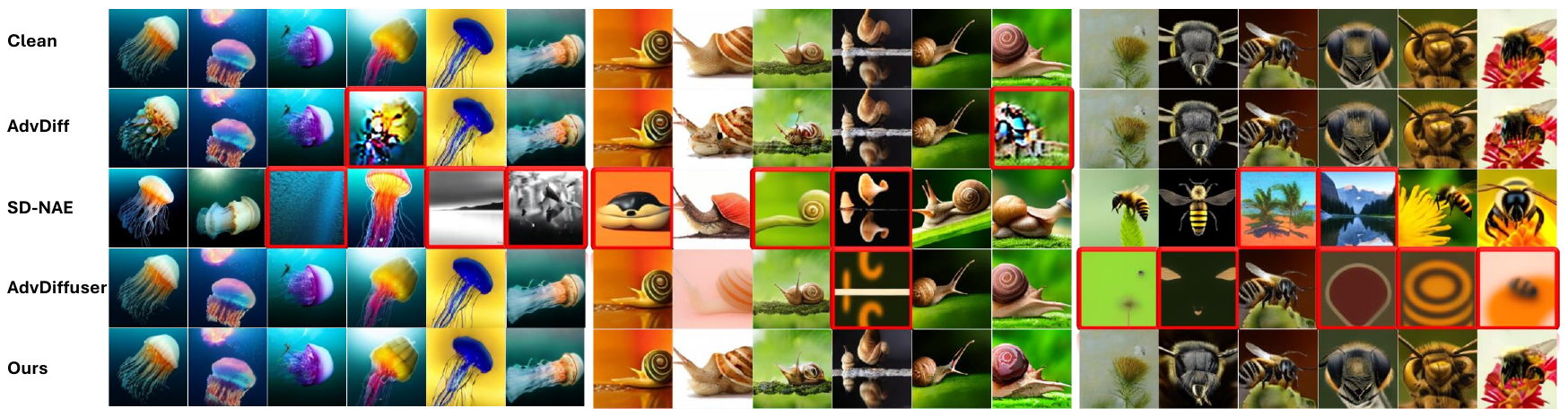}
    \caption{Note that most of existing adversarial attack methods can only work on given reference images, thus we generate the same reference NAEs from identical Gaussian noise and use basic class names as text prompts, facilitating a fair comparison across different attack strategies displayed in each column. The top row shows clean images produced by the stable diffusion model without adversarial perturbation, serving as a reference. Corrupted NAEs are outlined with red borders to ensure clear visual distinction.}
    \label{fig:experiment1}
\end{figure*}

\label{sec:experiment}
VENOM is designed to generate both NAEs and UAEs, and in this section, we evaluate its performance in both modes. We present comprehensive comparisons between VENOM and existing diffusion-based adversarial attack methods, focusing on both image quality and attack efficacy. Additionally, we conduct an ablation study to assess the impact of the adaptive control strategy and momentum on the adversarial guidance.

\subsection{Experimental Settings}
\textbf{Dataset and models:} We evaluate the NAE and UAE modes separately because they require different settings, and some works support only one of them. \cite{sd-nae, diffattack}. \textbf{NAE mode:} We generate NAEs from random Gaussian noise using text prompts derived from ImageNet labels. To ensure semantic alignment between generated images and their corresponding labels, we filter out ambiguous labels that are problematic for the Stable Diffusion model (e.g., ``drake", ``black widow"). Ambiguity is assessed by verifying whether a pretrained ResNet50 correctly classifies the generated image prompted by the label; labels leading to misclassification are discarded. This filtering process yields a subset of 466 usable labels from the original 1,000 ImageNet labels. (In the absence of adversarial perturbations, 72\% of clean images generated with filtered, unambiguous labels are accurately classified by a pretrained ResNet-50 model. However, without label filtering to remove ambiguous cases, the classification accuracy drops significantly to 40\%.) We generate five samples per selected class, resulting in a total of 2,330 NAEs. \textbf{UAE mode:} VENOM also generates UAEs based on reference images. We evaluate its performance on generating UAEs using the ImageNet-Compatible Dataset.\footnote{\url{https://github.com/cleverhans-lab/cleverhans/tree/master/cleverhans_v3.1.0/examples/nips17_adversarial_competition/dataset}} 
We utilize the Stable Diffusion model \cite{stable_diffusion} and DDIM \cite{ddim} sampling for generating both NAEs and UAEs. Refer to the supplementary material for more implementation details. 

\noindent\textbf{Evaluation metrics:} We evaluate the quality of UAEs and NAEs based on image quality and attack efficacy. Image quality is assessed using FID \cite{fid}, SSIM \cite{ssim}, LPIPS \cite{lpips}, Inception Score(IS) \cite{inceptionscore}, TReS score \cite{tres}, and CLIP score \cite{clipscore}. Attack efficacy is measured by the attack success rate (ASR) against both white box and black box models and defense methods such as Neural Representation Purifier(NRP) \cite{nrp}, Random Smoothing(RS) \cite{rs}, adversarial training \cite{madry_at} and DiffPure \cite{diffpure}.

\subsection{Synthesizing NAEs from Random Noise}
To date, three diffusion model-based adversarial attack methods have been proposed to synthesize Natural Adversarial Examples (NAEs) from random noise: SD-NAE \cite{sd-nae}, AdvDiffuser \cite{advdiffuser}, and AdvDiff \cite{advdiff}. These methods, however, utilize different diffusion models and inputs. Specifically, SD-NAE accepts text prompts and is based on Stable Diffusion \cite{stable_diffusion} with DDIM sampling \cite{ddim}, while AdvDiffuser and AdvDiff take class labels as input and are built upon Guided Diffusion \cite{guideddiffusion} with DDPM sampling and Latent Diffusion \cite{stable_diffusion} with DDIM sampling, respectively. Our proposed method, VENOM, employs text prompts and leverages Stable Diffusion with DDIM sampling.

To ensure a fair comparison under consistent conditions, we re-implement AdvDiff and AdvDiffuser using the same pretrained Stable Diffusion model and DDIM sampling. This standardization allows us to provide all adversarial attack methods with identical inputs and to rigorously evaluate the quality of the generated NAEs. Image quality is assessed using standard image quality evaluation metrics, and attack efficacy is evaluated based on the Attack Success Rate (ASR) against white-box models, black-box models, and various defense methods.

\noindent \textbf{Image quality assessment:} 
Most image quality evaluation metrics—such as FID, SSIM, and LPIPS—require reference images for comparison. Therefore, we generate clean images without adversarial guidance to serve as references. Table \ref{tab:image_quality_nae}
 reports the metric scores for NAEs generated by different methods, with arrows indicating whether higher or lower values are preferable. Our VENOM method achieves superior scores across FID, SSIM, LPIPS, and CLIP Score, demonstrating enhanced image quality that is structurally and perceptually closer to the originals and better aligned with textual descriptions than other methods. To provide qualitative insights beyond quantitative metrics, we display NAEs generated by various approaches in Figure \ref{fig:experiment1}. All images are generated from identical random noise inputs to facilitate direct visual comparison. We observe that other methods exhibit instability, introducing artifacts in certain samples: AdvDiff induces strong distortions, SD-NAE produces images with misaligned labels, and AdvDiffuser excessively denoises some examples. This instability suggests that their NAE generation capabilities are inconsistent, with some samples rendered satisfactorily while others are significantly corrupted due to the injection of adversarial guidance.
 
\begin{table}[ht]
\vspace{-18pt}
    \centering
    \caption{Image quality assessment of NAEs. The best result is bolded.}
    \scalebox{0.8}{
    \begin{tabular}{l|c|c|c|c|c}
        \toprule
        Method & FID $\downarrow$ & SSIM $\uparrow$ & LPIPS $\downarrow$ &IS $\uparrow$ & CLIP $\uparrow$ \\
        \midrule
        SD-NAE \cite{sd-nae}      &  27.78    &   0.2502   &  0.5601    & 42.27     &  0.8477    \\
        AdvDiffuser \cite{advdiffuser}&  21.34    &    0.7866  &  0.1608    & \textbf{44.40}     &   0.8765   \\
        AdvDiff \cite{advdiff}    &   34.25   &   0.8539   &  0.0763    &   30.20   & 0.8691     \\
        VENOM (Ours)      &   \textbf{14.49}   &  \textbf{0.8771}    & \textbf{0.0583}     &  41.12    &   \textbf{0.8789}   \\
        \bottomrule
    \end{tabular}}
    \label{tab:image_quality_nae}
    \vspace{-8pt}
\end{table}

\noindent \textbf{Attack efficacy:} We employ a pretrained ResNet50\cite{res50} as the victim model for all attack methods to generate NAEs. We evaluate the efficacy of different attacks by comparing their attack success rates (ASRs) on both white-box and black-box models. The white-box model is ResNet50 (Res50), while the black-box models include Inception-v3 (Inc-v3)\cite{inc-v3}, Vision Transformer (ViT)\cite{vit}, and MLP-Mixer Base (Mix-B)\cite{mlp}, representing convolutional neural network (CNN), transformers, and multi-layer perceptrons (MLPs) architectures, respectively. Additionally, we assess the robustness of various defense methods against these attacks, including adversarially trained ResNet50 (Res50-adv)\cite{res50adv}, NRP\cite{nrp}, RS\cite{rs}, and DiffPure\cite{diffpure}. Table \ref{tab:attack-efficacy-nae} summarizes the results of our attack efficacy evaluation. VENOM demonstrates the strongest performance in the white-box setting among all methods tested. However, its transferability to black-box models is markedly poor, as evidenced by the data in Table \ref{tab:attack-efficacy-nae}. While SD-NAE appears to achieve superior transferability and robustness against defense methods, a combined analysis of Table \ref{tab:attack-efficacy-nae} and Figure  \ref{fig:experiment1}  reveals that the reported high ASRs are misleading. Specifically, SD-NAE, AdvDiffuser, and AdvDiff suffer from instability in NAE generation; notably, SD-NAE and AdvDiffuser yield invalid NAEs in nearly half of the cases. A fair evaluation is infeasible without manual filtering of corrupted examples via human inspection, making white-box attack performance the only reliable metric.

\begin{table}[ht]
    \centering
    \caption{Attack success rates of NAEs against white-box (1st row), black-box models (2-4 rows), and various defense methods (5-8 rows). The best result is bolded.}
    \scalebox{0.72}{
    \begin{tabular}{l|c|c|c|c|c}
        \toprule
        Method & SD-NAE  & AdvDiff & AdvDiffuser & VENOM (ours)\\
        \midrule
        Res50 \cite{res50}   & 55.15 &  98.80  & 31.38 & \textbf{99.18}   \\
        \midrule
        Inc-V3 \cite{inc-v3} & 57.33    & \textbf{60.65}  & 41.12    & 50.39 \\
        Vit \cite{vit}   & \textbf{50.86}  & 42.45  &  33.05    &  34.98  \\
        Mix-B \cite{mlp} & \textbf{61.51} & 58.20&45.20 & 50.80 \\
        \midrule
        Res50-adv \cite{res50adv} &  53.40 & 39.02  &  36.87 & 34.68 \\
        NRP \cite{nrp} & 76.62  &  \textbf{85.63}    &66.61    &  85.46  \\
        RS  \cite{rs}   &  \textbf{55.54}   & 54.42 & 39.06 & 44.98 \\
        DiffPure \cite{diffpure} & 55.24 & 52.54  & 38.33 &   43.70  \\
        \bottomrule
    \end{tabular}
    }
    \label{tab:attack-efficacy-nae}
    \vspace{-8pt}
\end{table}

\subsection{Synthesizing UAEs from Reference Images}

\begin{figure}[ht]
    \centering
    \includegraphics[width=0.9\linewidth]{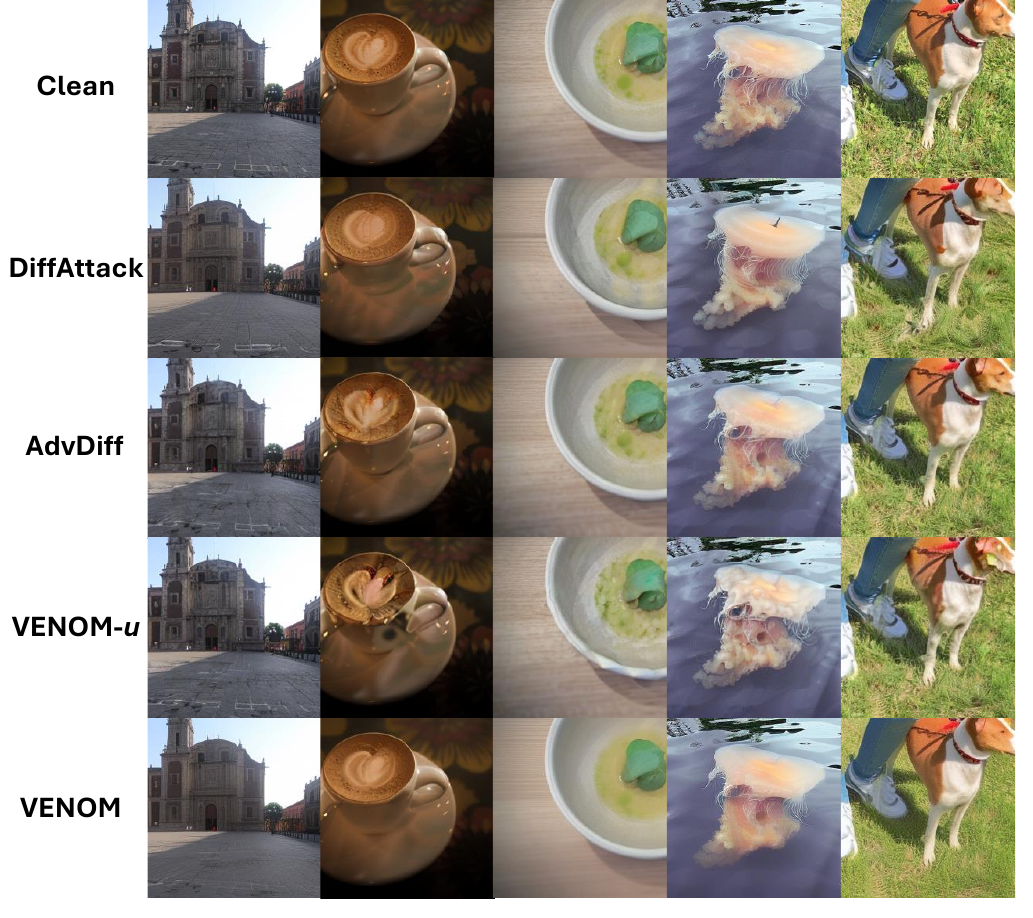}
    \caption{UAEs generated with different attack methods from reference clean images. Please zoom in to compare details.}
    \label{fig:uaes}
\end{figure}

\noindent \textbf{Attack efficacy:} We generate UAEs using existing images from the ImageNet-compatible dataset and evaluate the attack efficacy of different methods using the same metrics as for NAEs. The results are presented in Table \ref{tab:transferbility-uaes}, and Figure \ref{fig:uaes} showcases examples of the generated UAEs. A comparison between Figures \ref{fig:uaes} and \ref{fig:experiment1} reveals that the adversarial examples are visually almost identical to the original clean images across all attack methods, differing only in minor details. This high fidelity ensures the reliability of the attack efficacy evaluation metrics, as all UAEs are valid, contrasting with the potentially misleading results observed in Table \ref{tab:attack-efficacy-nae}. In Table \ref{tab:transferbility-uaes}, ``VENOM'' denotes the targeted attack version of our method, while ``VENOM-\textit{u}'' represents the untargeted variant. We include VENOM-\textit{u} for a fair comparison with DiffAttack \cite{diffattack}, which operates in an untargeted manner. The untargeted attack is implemented by setting the target label $y_a$ to the ground truth label $y$ and reversing the gradient sign in Eq. (\ref{eq:compute-gradient}). 

The untargeted version exhibits strong attack potency against all defense methods. DiffAttack demonstrates superior transferability on transformer and MLP based models but lower ASRs against the defense methods. Conversely, AdvDiff achieves the best performance in white-box attack but shows the poorest results in transferability and effectiveness against defense methods. We also evaluated the untargeted version of AdvDiff, which reportedly has good transferability; however, the image quality was significantly degraded, leading us to deem them invalid UAEs. Overall, our VENOM method exhibits a well-balanced performance across these evaluations, despite not being specifically designed for generating UAEs based on reference images.

\begin{table}[ht]
    \centering
    \caption{Attack success rates of UAEs against white-box (1st row), black-box models (2-4 rows), and various defense strategies (5-8 rows). The best result is bolded.}
    \scalebox{0.75}{
    \begin{tabular}{l|c|c|c|c}
        \toprule
        Method & DiffAttack & AdvDiff & VENOM & VENOM-\textit{u}\\
        \midrule
        Res50 \cite{res50}     & 94.30   & \textbf{99.80}    & 98.70  & 98.70 \\
        \midrule
        Inc-V3 \cite{inc-v3} &  70.80  & 36.52     &  45.80  &  \textbf{71.00}        \\
        Vit \cite{vit}    & \textbf{50.40}     & 12.83    &  17.20   & 40.60 \\
        Mix-B  \cite{mlp}    & \textbf{61.10}  & 29.24 &  36.00   &  59.40    \\
        \midrule
        Res50-adv \cite{res50adv} & 47.00 & 33.13  & 37.90 & \textbf{53.30} \\
        NRP \cite{nrp}     & 85.91 & 77.34  &  78.40  & \textbf{90.10}  \\
        RS \cite{rs} & 65.60     &  31.39   &  42.20   & \textbf{67.90}  \\
        DiffPure \cite{diffpure}  &  56.30  & 25.44 & 34.60 &  \textbf{60.10}\\
        \bottomrule
    \end{tabular}
}
    \label{tab:transferbility-uaes}
    \vspace{-8pt}
\end{table}

\noindent \textbf{Image quality assessment:} Figure \ref{fig:uaes} illustrates that UAEs generated by different methods exhibit relatively high imperceptibility. Table  \ref{tab:image_quality_uae} presents the scores from various image quality assessment metrics. Notably, the highest scores across different metrics are distributed among different attack methods, indicating that image quality evaluations vary depending on the assessment standards. The untargeted variant VENOM-\textit{u} attains slightly lower scores compared to DiffAttack, which can be attributed to its deviation from the real image distribution due to untargeted adversarial guidance. Nevertheless, the targeted version VENOM demonstrates strong overall performance(achieving the best or second-best scores across all metrics) by effectively balancing high perceptual similarity (LPIPS), structural consistency (SSIM), and superior image quality and diversity(TReS, FID and IS).

\begin{table}[ht]
\vspace{-12pt}
    \centering
    \caption{Image quality assessment of UAEs, the best result is bolded, and the second best is marked underlined.}
    \scalebox{0.8}{
    \begin{tabular}{l|c|c|c|c|c}
        \toprule
        Method & FID $\downarrow$ & SSIM $\uparrow$ & LPIPS$\downarrow$ & IS$\uparrow$ & TReS $\uparrow$ \\
        \midrule
        DiffAttack \cite{diffattack}    &    55.84  &   \textbf{0.7166}   &  0.1404    & 30.11     &  63.83    \\
        AdvDiff \cite{advdiff} & \textbf{34.52}     &  0.2273    &  0.7123     & \textbf{42.60}      &  60.63  \\
        
        VENOM-\textit{u}      &   62.15   &  0.6788    & 0.1856     &  22.72    &   61.04   \\
        VENOM    &  \underline{36.40}     & \underline{0.7151}    & \textbf{0.1399}     &  \underline{36.80}      &  \textbf{66.26}  \\
        \bottomrule
    \end{tabular}
    }
    \label{tab:image_quality_uae}
    \vspace{-8pt}
\end{table}

\subsection{Ablation Study}

\begin{figure}[ht]
\begin{minipage}[t]{.62\linewidth}

    \centering
    \captionof{table}{Ablation study evaluating the impact of the momentum (Mo) and adaptive control strategy (AS) modules.}
    \label{tab:ablation}
    \scalebox{0.70}{
    \begin{tabular}{cc|cccc}
        \toprule
        Mo & AS & FID$\downarrow$ & Res50 & DiffPure   \\
        \midrule
        \ding{55} & \ding{55} & 36.11 & 98.80  & \textbf{50.74}   \\
        \ding{52} & \ding{55}& 32.50 &\textbf{99.83} & 49.28 \\
        \ding{55} & \ding{52} & 15.09  & 98.89 & 43.70 \\
        \ding{52} & \ding{52}  &\textbf{14.49}  & 99.18 & 45.63 \\
        \bottomrule
    \end{tabular}
    }
       
  \end{minipage}
  \begin{minipage}[t]{.37\linewidth}
  
    \centering
    \captionof{figure}{$\beta \mathtt{\sim}$   FID}
    \label{fig:ablation}
    \includegraphics[width=\linewidth]{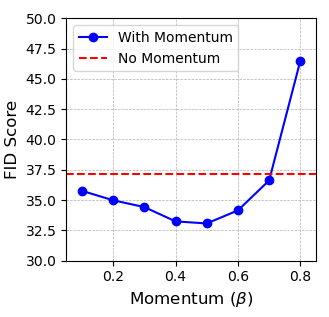}
    
  \end{minipage}
  \vspace{-8pt}
\end{figure}

Table \ref{tab:ablation} presents the effects of the momentum (Mo) and adaptive control strategy (AS) modules in VENOM. The adaptive control strategy primarily enhances image quality, as evidenced by improved FID scores; however, it reduces the ASR against white-box models and defensive mechanism. In contrast, the Momentum module contributes to image quality while compensating for the ASR loss introduced by the AS module. Together, Mo and AS modules achieve an optimal balance between image quality and attack effectiveness within the VENOM framework.

To determine the optimal hyperparameter $\beta$ for the Momentum module, we illustrate $\beta$ versus FID scores in Figure \ref{fig:ablation}, which shows that setting $\beta = 0.5$ yields the lowest FID score. The red dashed line in Figure \ref{fig:ablation} indicates the FID score of images generated without the Momentum module. Thus, we set $\beta = 0.5$ to maintain optimal image quality in the VENOM framework.

\subsection{Discussion\&Limitation} 

The UAEs and NAEs generated by VENOM achieve high white-box ASRs; however, their transferability to black-box models is lower compared to \cite{diffattack}, which focuses solely on generating UAEs from reference images. We argue that, \textbf{the high transferability from other attack methods is attributed to invalid NAEs} that the original image contents are completely distorted (see more in Supplementary Materials). Fairly measuring ASRs across black-box models and image content high-fidelity without extensive human intervention remains challenging.

\section{Conclusion}
\label{sec:conclusion}


In conclusion, we present VENOM, a novel framework for generating high-quality, text-driven UAEs with enhanced stability and attack efficacy. By incorporating adaptive control and momentum-enhanced gradients, VENOM effectively balances adversarial robustness and image realism while mitigating corruption. Unlike prior methods, it achieves stable perturbations, enabling consistent generation of both NAEs and UAEs from random noise or reference images. Experimental results highlight VENOM’s strong attack success rates and high image fidelity, solidifying its value as a robust tool for advancing adversarial research and enhancing deep learning security.
\newpage
{
    \small
    \bibliographystyle{ieeenat_fullname}
    \bibliography{stamo}
}


\end{document}